\documentclass[11pt]{article}

\usepackage[preprint]{acl}

\usepackage{times}
\usepackage{latexsym}
\usepackage{amsmath}

\usepackage[T1]{fontenc}
\usepackage[utf8]{inputenc}

\usepackage{enumitem}
\usepackage{microtype}

\usepackage{inconsolata}

\usepackage{graphicx}
\usepackage{bm}
\usepackage{amsmath}
\usepackage{amssymb}
\usepackage{spverbatim}
\usepackage{tcolorbox}
\usepackage{dsfont}
\usepackage{float}
\usepackage{graphicx}
\usepackage{subcaption}
\usepackage{booktabs} % for professional tables
\usepackage{multirow}
\usepackage{makecell}
\usepackage{soul}

\title{Second Guess: Detecting Uncertainty Through Abstention and Answer Stability in Small Language Models}

\author{Ashwath {Vaithinathan Aravindan} \\
  University of Southern California\\Los Angeles, 90007, California,\\United States of America \\
  \texttt{vaithina@usc.edu} \\\And
  Mayank Kejriwal \\
  Information Sciences Institute\\4676 Admiralty Way \#1001,\\Los Angeles, 90292, California,\\United States of America \\
  \texttt{kejriwal@isi.edu} \\}

\begin{document}
\maketitle
\begin{abstract}
Large language models often generate confident but incorrect answers rather than abstaining when uncertain. This problem is particularly acute for small language models (SLMs), where computational constraints and autonomous operation amplify the need for reliable uncertainty detection. We propose \textbf{Second Guess}, a lightweight, parameter-free prompting technique for abstention in multiple-choice question answering (MCQA) that is well-suited for SLMs. Our key empirical insight is that models which truly know an answer will select it consistently, while uncertain models exhibit unstable behavior when an ``I don't know'' option is added. Evaluated on four open models (2B-8B parameters) and four benchmarks, Second Guess achieves the highest composite risk improvement of 10.81\%. Notably, it maintains an 8\% composite risk improvement on fine-tuned models where entropy-based methods degrade, and improves most for lower-performing models. All code and results required to reproduce this work is available in \url{https://github.com/Mystic-Slice/second-guess}
\end{abstract}

\begin{figure}[!h]
\centering
\includegraphics[width=\linewidth]{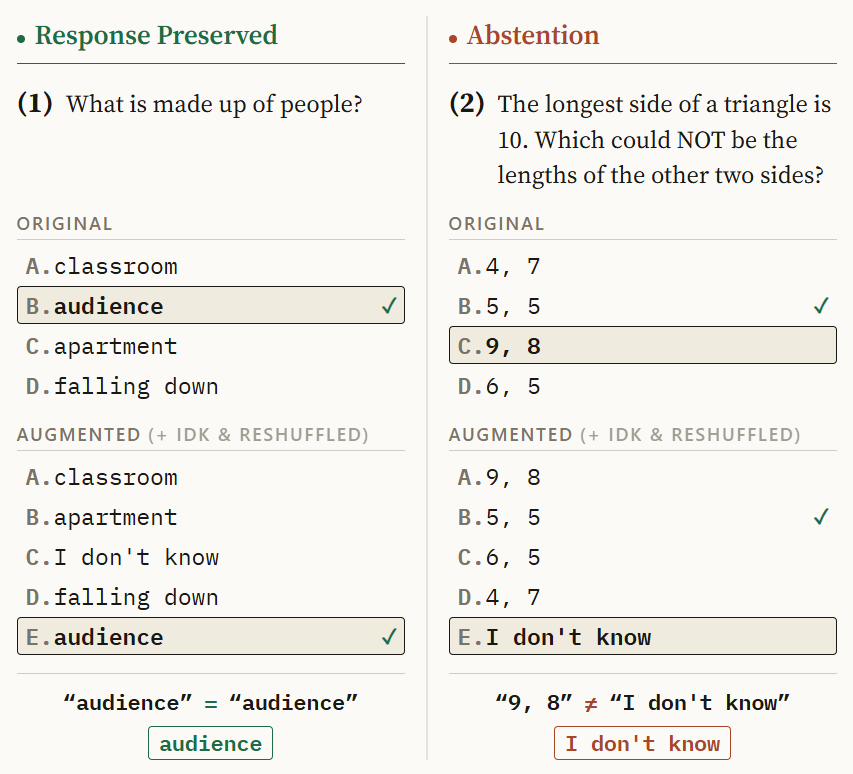}
\caption{Overview of the Second Guess technique. The model is prompted twice: once with the original options and once with an ``I don't know'' option added. If the model's answer changes between the two prompts, it is classified as an abstention; otherwise, the original answer is retained.}
\label{fig:second_guess_overview}
\end{figure}

\section{Introduction}
\label{sec:intro}

While large language models (LLMs) often generate confident, but incorrect, answers rather than acknowledging uncertainty and abstain, the challenge is even more acute for small language models (SLMs), denoted here as transformer-based generative models with fewer than 10B parameters. SLMs are important in many contexts, including deployment on edge and mobile devices~\cite{lu2024small}, where computational constraints can amplify the importance of reliable uncertainty detection, and autonomous operation precedes the availability of human oversight~\cite{tomani2024uncertainty}.

Recent work addresses abstention through diverse methodological approaches, each with distinct constraints. \textit{Training-based} methods~\citep{zhu2025craft,kanber2025abstain} leverage ``I don't know'' labels or dedicated tokens during fine-tuning, though they require direct access to model weights. \textit{Prompt-based} methods~\citep{zong2026icalm,kadavath2022language,ren2026uabench} attempt to elicit uncertainty through explicit queries, but their effectiveness depends on the assumption that models accurately self-report their confidence. \textit{Sampling-based} methods~\citep{cole2023selectively,wang2022selfc} achieve robustness through consensus across multiple model outputs, at the cost of substantial computational overhead.

In this paper, we propose \textbf{Second Guess}, a novel parameter-free prompting technique that operates under the key insight that models exhibiting stable knowledge will provide consistent answers \textit{regardless} of available options (Figure \ref{fig:second_guess_overview}). The method prompts each model twice per question: once with the standard choice set and again with an ``I don't know'' option added. We treat answer discrepancies between these two prompts as an abstention signal, while unchanged responses are retained as confident predictions. This approach demands only two forward passes per question, making it applicable even to black-box SLMs without access to (internal) model parameters or uncertainty scores.

We evaluate Second Guess on four SLMs (2B–8B parameters) and four MCQ benchmarks, and contribute: (i) a parameter-free abstention technique with only 2 queries and no access to internal state of the model; (ii) comprehensive evaluation showing Second Guess achieves highest composite risk improvements while maintaining robustness to fine-tuning; and (iii) empirical insights demonstrating that improvements stem from detecting unstable answer switching and showing the method is best suited for challenging tasks and lower-performing models.

\section{Related Work}
\label{sec:related_work}
Abstention, the ability to decline answering a question when uncertain, is critical for deployed language models, as incorrect predictions often cause greater harm than no prediction at all in safety critical domains like healthcare. Existing approaches include training-based methods that require weight access and retraining~\citep{zhu2025craft,kanber2025abstain} and prompt-based methods that rely on self-reported uncertainty~\citep{zong2026icalm,kadavath2022language}. Consistency-based approaches measure behavioral stability across perturbations~\citep{cole2023selectively}, but existing work uses unstructured stochastic sampling requiring many forward passes. While these approaches are well established, achieving reliable abstention in LLMs remains a challenge~\citep{kirichenko2026abstentionbench,wen2025know}. 

Small language models (SLMs), typically fewer than 10 billion parameters, increasingly deployed on edge and mobile devices where latency and privacy constraints preclude API calls~\citep{lu2024small,wang2025comprehensive}. Calibration generally improves with model scale~\citep{kalai2024calibrated,elhady2025wicked}: smaller models exhibit greater overconfidence and susceptibility to perturbations, thereby more in need of reliability harness around them~\citep{ aravindan2026fragile}. Crucially, SLMs are often deployed autonomously in embedded systems where incorrect answers carry direct consequences, yet sampling-based consistency methods~\citep{cole2023selectively} are computationally prohibitive. 

% Our approach bridges these threads with a simple, lightweight, parameter-free procedure. It uses option augmentation as a \emph{structured perturbation} whose effect reveals the model's confidence: a model that truly knows the answer will select it regardless of whether an IDK option is available, while a model that is guessing will be destabilized by the new option. To the best of our knowledge, no prior work has proposed or evaluated this paired-prompt consistency mechanism as an abstention strategy. 

\section{Methodology}
\label{sec:methodology}

\subsection{Second Guess}
Consider a question $Q$ and its four corresponding answer choices $\mathcal{C} = \{c_1, c_2, c_3, c_4\}$ presented in a zero-shot prompt\footnote{We reproduce a standard multiple-choice prompt template in Figure~\ref{fig:prompt_example} in Appendix~\ref{sec:appendix_prompt}, demonstrating the structured format used in our experiments.}. In the first stage, the model's response $R_1$ is extracted from the original prompt. In the second stage, an abstention option "I don't know" is inserted at a random position to $\mathcal{C}$ as the fifth option, and the model's response $R_2$ to the augmented prompt is extracted. If the model's answer remains consistent between stages ($R_1 = R_2$), the original answer choice is retained. Conversely, if the response changes ($R_1 \neq R_2$), regardless of what it changes to, the model's output is classified as an abstention response "I don't know".

Formally, the final model answer $R$ is derived as follows:
\begin{equation}
R = \begin{cases}
R_1 & \text{if } R_1 = R_2 \\
\text{``I don't know''} & \text{if } R_1 \neq R_2
\end{cases}
\end{equation}

\subsection{Experimental Setup}
\subsubsection{Models and Datasets}

We evaluate four instruction-tuned language models: Mistral-7B-Instruct-v0.3~\cite{Jiang2023Mistral7}, Llama-3.1-8B-Instruct~\cite{grattafiori2024llama}, IBM Granite-3.3-2B-Instruct\footnote{\url{https://huggingface.co/ibm-granite/granite-3.3-2b-instruct}}, and Qwen3-4B-Instruct-2507~\cite{yang2025qwen3}. More information about the models used is provided in Appendix A (Table~\ref{tab:models}). We evaluate model behavior across four multiple-choice question-answering datasets: CommonsenseQA~\cite{talmor-etal-2019-commonsenseqa}, QASC~\cite{allenai:qasc}, MMLU-Pro~\cite{wang2024mmlu}, and SuperGPQA~\cite{du2025supergpqa}. For each model and dataset, we sample 100 random questions and normalize them to contain exactly four options per question. A detailed summary of the datasets is provided in Appendix~\ref{sec:appendix_prompt} (Table~\ref{tab:datasets}).

\subsubsection{Baselines}

Because our approach is lightweight and requires only model outputs, we select baselines that are similarly lightweight and practical. Therefore complex and sampling based approaches like proposed by ~\cite{cole2023selectively} are not evaluated against. The baselines evaluated are:
\begin{enumerate}[nosep]
  \item \textbf{Original Prompt only}: The standard multiple-choice options $\mathcal{C} = \{c_1, c_2, c_3, c_4\}$ without augmentation. The options are shuffled randomly to avoid bias due to ordering.

  \item \textbf{Augmented Prompt only}: A single-stage variant where the abstention option ``I don't know'' is added at a random position to the original options from the outset, forming $\mathcal{C}' = \{c_1, c_2, c_3, c_4, \text{``I don't know''}\}$. This baseline directly exposes the model to the possibility of abstaining in a single pass. 

  \item \textbf{Self-Evaluation}~\cite{kadavath2022language}: A two-step prompting strategy adapted for abstention. In the first stage, the model generates an answer; in the second stage, it is presented with this answer and asked to verify its correctness, abstaining when it doesn't accept the proposed answer.

  \item \textbf{Entropy-based Uncertainty Estimation and Thresholding}~\cite{malinin2020uncertainty}: A post-hoc baseline where responses with output token entropy exceeding the mean plus one standard deviation of the entropy distribution across the entire dataset are automatically converted to abstentions; otherwise, the model's response is preserved. See Figure~\ref{fig:entropy_distribution} in Appendix~\ref{sec:entropy_thresholding_baseline} for more details. 
\end{enumerate}

\subsubsection{Fine-tuning with Low-Rank Adaptation}

To understand whether the advantage of the two-stage prompting technique persists when models are highly capable in MCQA task, we fine-tune two models (Qwen and Llama) using low-rank adaptation (LoRA)~\cite{hu2022lora}. Details regarding the hyperparameters used for model training are provided in Appendix~\ref{sec:hyperparams}.

\subsubsection{Evaluation Metrics}
We employ three metrics to evaluate model performance. We define the following notation: $N$ = total number of questions, $N_c$ = count of correct answers, $N_i$ = count of incorrect answers, $N_a$ = count of abstentions and $N_{ca}$ = count of abstentions that were correct answers in the original prompt. The three evaluation metrics are defined as follows:
\begin{enumerate}[nosep]
  \item \textbf{Precision} The proportion of non-abstained questions answered correctly. It is computed as: $\frac{N_c}{N_c + N_i} \times 100$
  \item \textbf{Error Rate} The proportion of incorrect answers relative to the total number of questions, computed as: $\frac{N_i}{N} \times 100$
  \item \textbf{Composite Risk} The proportion of problematic outcomes, combining both incorrect answers and cases where the model abstains when it provided a correct response in the original prompt~\cite{composite_risk}, computed as: $\frac{N_i + N_{ca}}{N} \times 100$
\end{enumerate}

\begin{table*}[h]
\centering
\caption{Average performance of each method across all base models and datasets. Each cell shows the mean and standard deviation of metric value followed by the averaged  percentage change versus the \textit{Original} baseline.}
\label{tab:main_summary_base}
\resizebox{\textwidth}{!}{
\begin{tabular}{lrrr}
\toprule
Method & Precision $\uparrow$ & Error Rate $\downarrow$ & Composite Risk $\downarrow$ \\
\midrule
Original & 59.75 $\pm$ 16.72 & 40.25 $\pm$ 16.72 & 40.25 $\pm$ 16.72 \\
Augmented & 61.04 $\pm$ 16.77 (+1.29) & 38.00 $\pm$ 15.99 (-2.25) & 38.00 $\pm$ 15.99 (-2.25) \\
Self-Evaluation & 62.10 $\pm$ 17.31 (+2.35) & 32.56 $\pm$ 15.18 (-7.69) & 32.56 $\pm$ 15.18 (-7.69) \\
Entropy Thresholding (Original) & 63.30 $\pm$ 18.03 (+3.55) & 32.00 $\pm$ 16.18 (-8.25) & 36.56 $\pm$ 15.34 (-3.69) \\
Entropy Thresholding (Augmented) & 63.49 $\pm$ 18.54 (+3.74) & 29.88 $\pm$ 14.78 (-10.38) & 34.69 $\pm$ 14.26 (-5.56) \\
\midrule
Second Guess (Ours*) & \textbf{68.40 $\pm$ 14.93 (+8.65)} & \textbf{20.12 $\pm$ 5.32 (-20.12)} & \textbf{29.44 $\pm$ 8.66 (-10.81)} \\
\bottomrule
\end{tabular}
}
\end{table*}

\section{Results}
\textbf{Finding 1: Second Guess provides a significant improvement in Composite Risk across models and datasets.}
Table~\ref{tab:main_summary_base} presents the average performance of different methods across all base models and datasets. The Augmented prompt only baseline demonstrates minimal improvement, with changes of $2.25\%$ in Composite Risk over the Original prompt. Entropy Thresholding shows moderate improvement (~3-6\% drop), especially when using the Augmented prompt. Self-Evaluation proves to be the best baseline with $7.69\%$ drop in Composite Risk. Whereas, Second Guess improves performance across all metrics, achieving particularly strong gains in Composite Risk with a $10.81\%$ drop, while using only 2 model queries, just as much as Self-Evaluation. Table~\ref{tab:detailed_second_guess_(ours*)} in Appendix~\ref{sec:appendix_baseline} presents a detailed breakdown across all model-dataset combinations.

\textbf{Finding 2: Second Guess also works on finetuned models while other baselines degrade}
Even for models that are already effective at the task, Second Guess delivers an 8\% improvement in Composite Risk, with Self-Evaluation also keeping up, whereas other methods achieve improvements of roughly 0-2.5\%. Notably, Entropy Thresholding loses out on finetuned models since with finetuning, the confidence of the models on that dataset increases and token entropy is no longer very informative on the models' uncertainty. Second Guess continues to offer almost double the reduction in Error Rate compared to Self-Evaluation while almost matching the improvement in Precision. In Appendix~\ref{sec:finetuned_results}, Table~\ref{tab:main_summary_finetuned} shows comparative results for fine-tuned models averaged across all datasets against baselines, while Table~\ref{tab:base_vs_finetuned} compares the results against the corresponding base models.

\begin{figure}[!h]
\centering
\includegraphics[width=\linewidth]{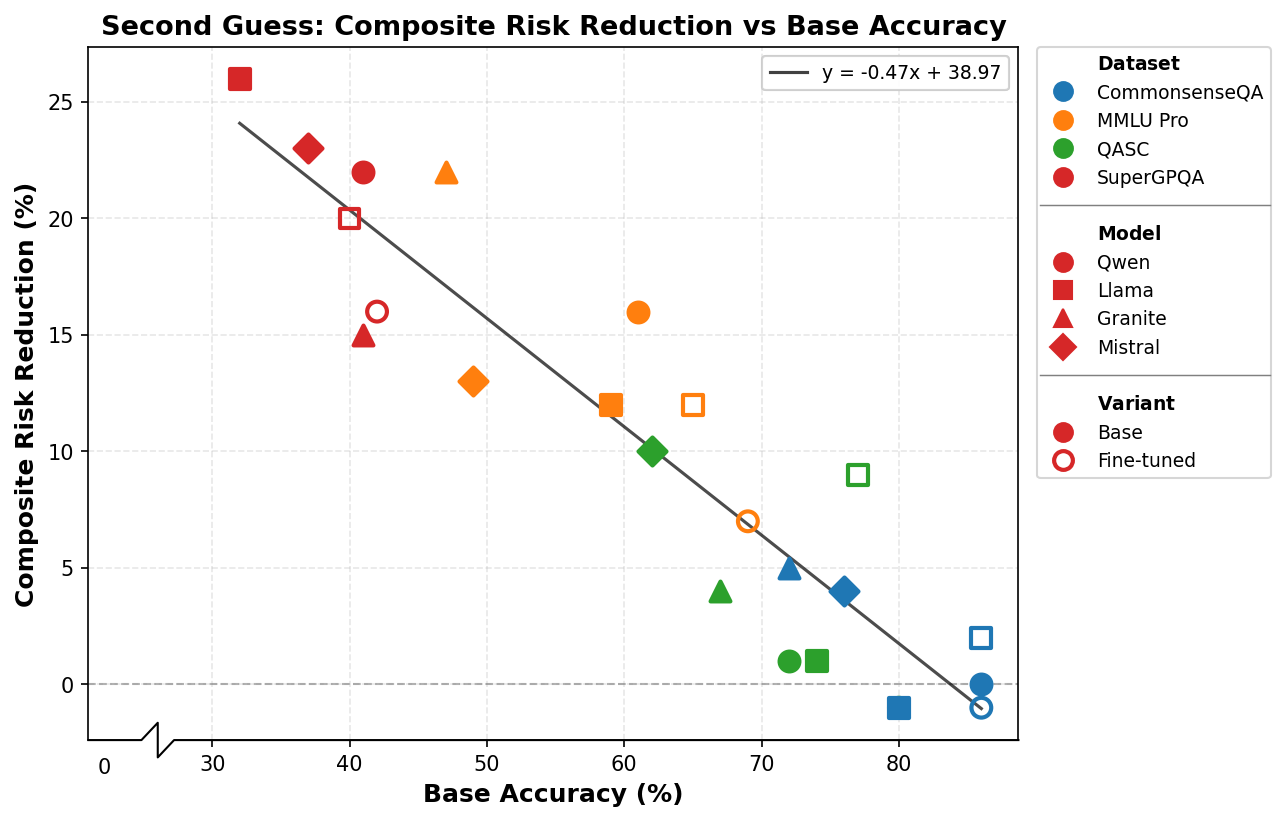}
\caption{Composite Risk Reduction vs. Base Accuracy. The scatter plot shows the relationship between a model's base accuracy and the improvement achieved by Second Guess across all model-dataset combinations. The negative linear trend (fitted line: $y = -0.47x + 38.97$) demonstrates that models with lower base accuracy benefit more from our approach.}
% , 
% with reductions exceeding 20\% for the lowest-performing models and diminishing gains as base accuracy increases.}
\label{fig:accuracy_improvement}
\end{figure}
\textbf{Finding 3: There is an inverse relationship between improvement in Composite Risk and the model's base performance}
Figure~\ref{fig:accuracy_improvement} reveals a clear inverse relationship between a model's base accuracy and the improvement in Composite Risk delivered by Second Guess. Specifically, models with lower base accuracy exhibit substantially higher Composite Risk reductions, while those with higher base accuracy see more modest improvements. The fitted line ($y = -0.47x + 38.97$) demonstrates that for every 1\% increase in base accuracy, the expected composite risk reduction decreases by approximately 0.47 percentage points. As also seen in Table~\ref{tab:detailed_second_guess_(ours*)} in Appendix~\ref{sec:appendix_baseline}, the models with the lowest base accuracy—such as Llama on SuperGPQA (base accuracy $\approx 32\%$)—benefit the most, achieving composite risk reductions exceeding 25\%. Conversely, models already performing well, such as Qwen on CommonsenseQA (base accuracy $\approx 85\%$), see no improvement. This observation suggests that Second Guess is more valuable for use in challenging domains, where model accuracy is lower and error reduction is most critical.

\textbf{Finding 4: Composite Risk gains primarily stem from option switching rather than "I don't know" selection}
Table~\ref{tab:change_breakdown_combined} in Appendix~\ref{sec:change_breakdown_combined} shows the distribution of models' choices in the Augmented prompt versus the Original prompt. The models rarely selects the "I don't know" option and instead tends to switch between wrong answers. For example, the Qwen model on MMLU Pro, switches only 2 of its incorrect answers in the Original Prompt to "I don't know" in the Augmented Prompt, while switching 21 of those to some other choices in the latter prompt, showing that most of the error rate reduction came from capturing this option switching behaviour rather than explicit abstention by the model. This pattern generalizes across all models and datasets, with multiple settings where models chose "I don't know" zero times.

\section{Discussion}

The experiments show that Second Guess can be effective in improving useful abstention and is also practical due to minimal computation requirements. Our evaluation spans a broad set of question types involving both reasoning and knowledge retrieval (Table~\ref{tab:datasets}), suggesting general applicability. Second Guess is also a  parameter-free method, requiring no tuning or hyperparameter optimization across different models or domains, making it accessible.

Interestingly, Finding 3 suggests that finetuned versions of models should benefit less from Second Guess compared to their base counterparts. For the Qwen model, finetuning does result in a reduction in the Composite Risk improvement observed (-9.75\% base vs -5.25\% finetuned). But for the Llama model, there is actually a modest increase in the Composite Risk improvement after finetuning (-9.50\% base vs -10.75\% finetuned) (see Table~\ref{tab:base_vs_finetuned} in Appendix~\ref{sec:finetuned_results}). Hence, behavioral uncertainty exploited by Second Guess is likely unaffected by finetuning, and may be inherent to the model itself, possibly connected to knowledge gained during pre-training~\cite{zhou2023lima}.

\section{Limitations}
This study has several important limitations that should be considered when interpreting the results. First, the study focusses on smaller language models with fewer than 10 billion parameters. The analysis in Finding 3 reveals an inverse relationship between model capability and improvement magnitude: as base model performance increases, the gains from Second Guess diminish. This pattern raises an important question about scalability to frontier-scale models, which already achieve high performance on many benchmarks. For frontier models, Second Guess may primarily benefit only in specialized domains requiring niche knowledge not already in the training data or tasks that remain challenging despite their scale. However, this limitation should not diminish the significance of improving the reliability of smaller models. Smaller language models remain critical in many real-world applications~\cite{belcak2025small}, particularly in scenarios with stringent latency, computational, or privacy constraints. In such settings, Second Guess demonstrates substantial value by consistently improving performance and reliability without incurring massive additional inference costs.

Furthermore, this study evaluates models without Chain-of-Thought (CoT) reasoning~\cite{wei2022chain}, which has become standard practice for extracting maximum performance from modern language models. Although our evaluation shows performance variation across models (e.g., Base Qwen achieving 86\% on CommonsenseQA while Llama scores only 32\% on SuperGPQA, as detailed in Table~\ref{tab:appendix_original} in Appendix~\ref{sec:appendix_baseline}), indicating that Second Guess benefits models across performance levels, the absence of CoT likely underestimates the baseline capabilities of evaluated models. Future work should investigate how Second Guess interacts with advanced prompting strategies to establish its utility when models operate at closer to their maximum potential.

\bibliography{custom}

\newpage
\appendix

\section*{Appendix Contents}
\begin{itemize}
  \item[\textbf{A}] Prompt, Models and Datasets \dotfill \pageref{sec:appendix_prompt}
  \item[\textbf{B}] LoRA Hyperparameter Details \dotfill \pageref{sec:hyperparams}
  \item[\textbf{C}] Entropy Thresholding Baseline \dotfill \pageref{sec:entropy_thresholding_baseline}
  \item[\textbf{D}] Results of Finetuned Models \dotfill \pageref{sec:finetuned_results}
  \item[\textbf{E}] Results for Baseline Methods \dotfill \pageref{sec:appendix_baseline}
  \item[\textbf{F}] Second Guess: Change Breakdown for Remaining (Dataset, Model) Pairs \dotfill \pageref{sec:change_breakdown_combined}
\end{itemize}

\newpage
\section{Prompt, Models and Datasets}\label{sec:appendix_prompt}
Figure~\ref{fig:prompt_example} shows the prompt template used for the multiple-choice question answering used in experiments. Tables~\ref{tab:models} and \ref{tab:datasets} provide more information about the models and datasets used respectively.

\begin{figure*}[!h]
\centering
\begin{tcolorbox}[colback=white,colframe=black,width=\linewidth]
You are given a question and some options. Output the correct option letter only and nothing else.

\textless question\textgreater 

For the two linear equations 2 * x + 3 * y = 10 and 4 * x + 4 * y = 12 with variables x and y. Use cramer's rule to solve these two variables.

\textless /question\textgreater

\textless options\textgreater

(A) [4, 1]

(B) [-2, 6]

(C) [3, 2]

(D) [-1, 4]

(E) I don't know

\textless /options\textgreater

The correct option is: (
\end{tcolorbox}
\caption{Prompt template used in the MCQA tasks.}
\label{fig:prompt_example}
\end{figure*}

\begin{table*}[!h]
\caption{List of models evaluated in this study. Release dates are approximate and sourced from Hugging Face and arXiv.}
\label{tab:models}
\centering
\resizebox{\textwidth}{!}{
\begin{tabular}{lclp{4cm}}
\toprule
\textbf{Model} & \textbf{Parameter Count} & \textbf{Release Date} & \textbf{License} \\
\midrule
\href{https://huggingface.co/mistralai/Mistral-7B-Instruct-v0.3}{mistralai/Mistral-7B-Instruct-v0.3} & 7B & May 22, 2024 & Apache 2.0\\
\href{https://huggingface.co/meta-llama/Llama-3.1-8B-Instruct}{meta-llama/Llama-3.1-8B-Instruct} & 8B & July 23, 2024 & Llama 3.1 Community License Agreement\\
% \href{https://huggingface.co/google/gemma-3-4b-it}{google/gemma-3-4b-it} & 4B & March 11, 2025\\s
\href{https://huggingface.co/ibm-granite/granite-3.3-2b-instruct}{ibm-granite/granite-3.3-2b-instruct} & 2B & April 16, 2025 & Apache 2.0\\
\href{https://huggingface.co/Qwen/Qwen3-4B-Instruct-2507}{Qwen/Qwen3-4B-Instruct-2507} & 4B & August 5, 2025 & Apache 2.0\\
\bottomrule
\end{tabular}
}
\end{table*}

\begin{table*}[!h]
\caption{Summary of datasets used in this study. Release dates are approximate and sourced from Hugging Face and arXiv.}
\label{tab:datasets}
\centering
\resizebox{\textwidth}{!}{
\begin{tabular}{lp{6cm}lp{4cm}}
\toprule
\textbf{Dataset} & \textbf{Description} & \textbf{Release Date} & \textbf{License} \\
\midrule
\textbf{\href{https://huggingface.co/datasets/tau/commonsense_qa}{CommonsenseQA}} & Benchmark for evaluating commonsense reasoning capabilities through multiple-choice questions. & November 2, 2018 & MIT\\
\textbf{\href{https://huggingface.co/datasets/allenai/qasc}{QASC}} & Question-answering dataset requiring compositional reasoning and multi-step inference across multiple facts. & October 21, 2019 & Creative Commons Attribution 4.0\\

\textbf{\href{https://huggingface.co/datasets/TIGER-Lab/MMLU-Pro}{MMLU-Pro}} & Enhanced MMLU benchmark covering STEM, social sciences, and humanities with challenging multiple-choice questions. & May 8, 2024 & MIT\\
\textbf{\href{https://huggingface.co/datasets/m-a-p/SuperGPQA}{SuperGPQA}} & Graduate-level question-answering benchmark covering advanced concepts in physics, chemistry, and biology. & February 20, 2025 & Open Data Commons License Attribution\\
\bottomrule
\end{tabular}
}
\end{table*}

\section{LoRA Hyperparameter Details}\label{sec:hyperparams}

\paragraph{Training Hyperparameters}
Hyperparameters for all LoRA model trainings are set as follows:
\begin{itemize}
  \item Learning rate: $2 \times 10^{-4}$
  \item Maximum prompt length: $1024$
  \item Batch size: $64$
  \item LoRA rank: $32$
\end{itemize}

\paragraph{Implementation}
We use the Tinker library for implementing LoRA fine-tuning.

\paragraph{Dataset-specific Sample Counts}
The number of samples used for fine-tuning varies by dataset:
\begin{itemize}
  \item CommonsenseQA: $9000$ samples
  \item MMLU-Pro: $10000$ samples
  \item QASC: $8000$ samples
  \item SuperGPQA: $10000$ samples
\end{itemize}

\section{Entropy Thresholding Baseline}
\begin{figure*}[!h]
\centering
\includegraphics[width=0.9\textwidth]{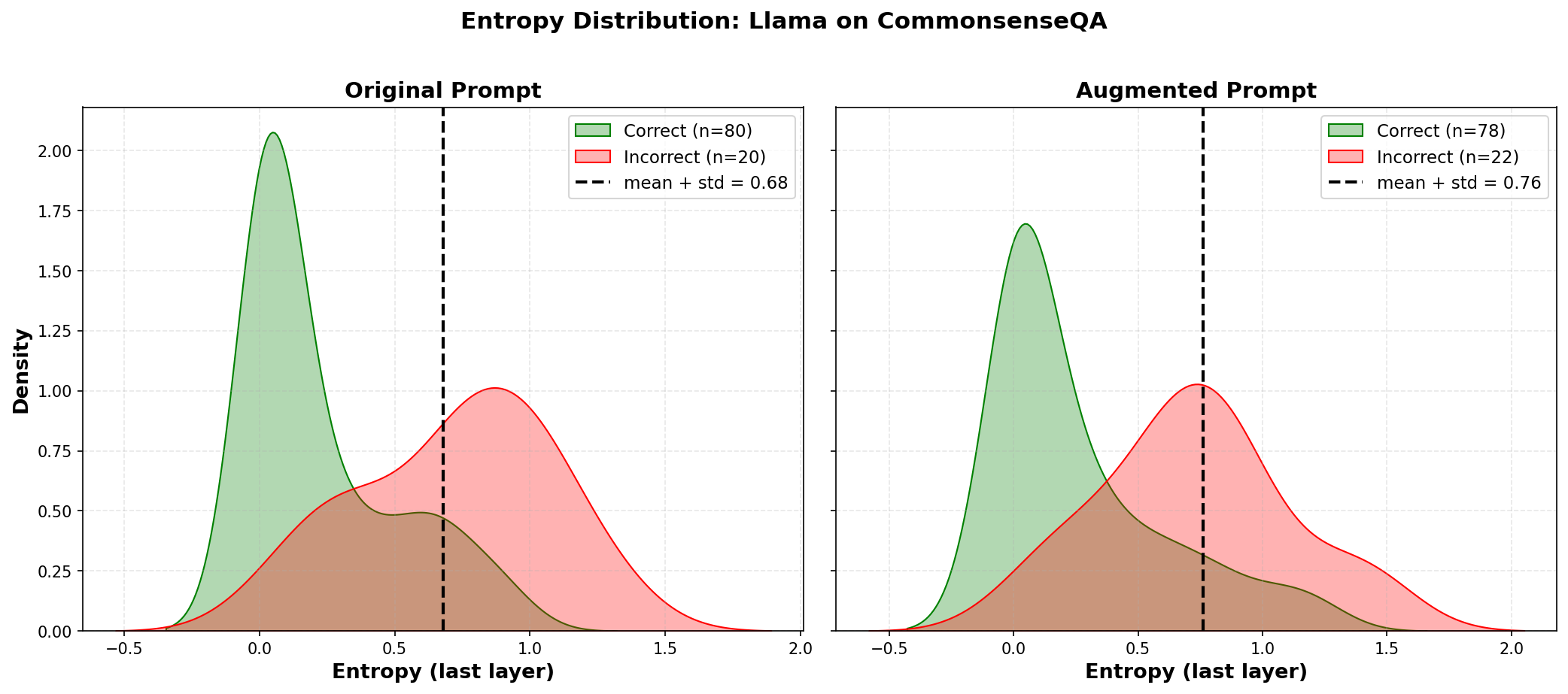}
\caption{Entropy distribution for the Llama model on CommonsenseQA dataset comparing original and augmented prompts. The histogram shows the distribution of entropy values across correct predictions (green) and incorrect predictions (red). The dashed line indicates the threshold (mean + standard deviation) used to determine whether predictions should be abstained from. This visualization demonstrates how entropy-based thresholding leverages the overlap between the two distributions to selectively abstain from low-confidence predictions.}
\label{fig:entropy_distribution}
\end{figure*}

The threshold value of mean + standard deviation was selected from a pool of statistical candidates for the thresholds ({mean, mean + std/2, mean + std, mean + 2*std}). Figure~\ref{fig:entropy_distribution} shows how separation of responses for abstention works on the overall entropy distribution.
\label{sec:entropy_thresholding_baseline}

\clearpage
\section{Results of Finetuned models}
\label{sec:finetuned_results}
Tables~\ref{tab:main_summary_finetuned} compares the performance of Second Guess against the baselines for the finetuned Qwen and Llama models in isolation. Table~\ref{tab:base_vs_finetuned} shows how Second Guess helps the finetuned Qwen and Llama models in comparison to their base counterparts.

\begin{table*}[h]
\centering
\caption{Average performance for fine-tuned models, each evaluated only on its fine-tuning dataset. Cells show the mean and standard deviation of metric value followed by the averaged  percentage change vs the Original baseline.}
\label{tab:main_summary_finetuned}
\resizebox{\textwidth}{!}{
\begin{tabular}{lrrr}
\toprule
Method & Precision $\uparrow$ & Error Rate $\downarrow$ & Composite Risk $\downarrow$ \\
\midrule
Original & 68.12 $\pm$ 18.29 & 31.88 $\pm$ 18.29 & 31.88 $\pm$ 18.29 \\
Augmented & 70.19 $\pm$ 17.71 (+2.06) & 29.38 $\pm$ 17.05 (-2.50) & 29.38 $\pm$ 17.05 (-2.50) \\
Self-Evaluation & \textbf{75.59 $\pm$ 18.23 (+7.46)} & 24.00 $\pm$ 18.13 (-7.88) & 24.00 $\pm$ 18.13 (-7.88) \\
Entropy Thresholding (Original) & 73.40 $\pm$ 21.31 (+5.27) & 24.50 $\pm$ 22.23 (-7.38) & 31.25 $\pm$ 17.61 (-0.63) \\
Entropy Thresholding (Augmented) & 74.20 $\pm$ 19.96 (+6.08) & 22.38 $\pm$ 18.75 (-9.50) & 30.50 $\pm$ 14.31 (-1.38) \\
\midrule
Second Guess (Ours*) & 75.28 $\pm$ 16.05 (+7.16) & \textbf{17.38 $\pm$ 6.99 (-14.50)} & \textbf{23.88 $\pm$ 11.43 (-8.00)} \\
\bottomrule
\end{tabular}
}
\end{table*}

\begin{table*}[h]
\centering
\small
\caption{Second Guess average performance comparing base and fine-tuned variants of Qwen and Llama, restricted to the datasets where both variants are available. Each cell shows the mean and standard deviation of metric value followed by the averaged percentage change vs the Original baseline.}
\label{tab:base_vs_finetuned}
\begin{tabular}{lrrr}
\toprule
Variant & Precision $\uparrow$ & Error Rate $\downarrow$ & Composite Risk $\downarrow$ \\
\midrule
Qwen (Base) & 73.94 $\pm$ 14.45 (+8.94) & 17.75 $\pm$ 5.56 (-17.25) & 25.25 $\pm$ 9.54 (-9.75) \\
Qwen (FT) & 74.97 $\pm$ 17.02 (+5.72) & 17.50 $\pm$ 6.81 (-13.25) & 25.50 $\pm$ 11.62 (-5.25) \\
Llama (Base) & 68.22 $\pm$ 18.87 (+6.97) & 21.75 $\pm$ 7.27 (-17.00) & 29.25 $\pm$ 9.11 (-9.50) \\
Llama (FT) & 75.59 $\pm$ 17.64 (+8.59) & 17.25 $\pm$ 8.22 (-15.75) & 22.25 $\pm$ 12.76 (-10.75) \\
\bottomrule
\end{tabular}
\end{table*}

\newpage
\section{Results for Baseline Methods}\label{sec:appendix_baseline}
Tables~\ref{tab:appendix_original}, \ref{tab:appendix_augmented}, \ref{tab:appendix_self_evaluation}, \ref{tab:appendix_entropy_thresholding_(original)}, \ref{tab:appendix_entropy_thresholding_(augmented)}, and \ref{tab:detailed_second_guess_(ours*)} contain per-(dataset, model) results for Original, Augmented, Self-Evaluation, Entropy Thresholding (Original), Entropy Thresholding (Augmented), and Second Guess (Ours*) respectively

\begin{table*}[!h]
\centering
\small
\caption{Original baseline per-(dataset, model) results across all models (base and fine-tuned). Raw metric values.}
\label{tab:appendix_original}
\begin{tabular}{llrrr}
\toprule
Dataset & Model & Precision $\uparrow$ & Error Rate $\downarrow$ & Composite Risk $\downarrow$ \\
\midrule
CommonsenseQA & Granite & 72.00 & 28.00 & 28.00 \\
CommonsenseQA & Llama & 80.00 & 20.00 & 20.00 \\
CommonsenseQA & Llama-FT & 86.00 & 14.00 & 14.00 \\
CommonsenseQA & Mistral & 76.00 & 24.00 & 24.00 \\
CommonsenseQA & Qwen & 86.00 & 14.00 & 14.00 \\
CommonsenseQA & Qwen-FT & 86.00 & 14.00 & 14.00 \\
MMLU Pro & Granite & 47.00 & 53.00 & 53.00 \\
MMLU Pro & Llama & 59.00 & 41.00 & 41.00 \\
MMLU Pro & Llama-FT & 65.00 & 35.00 & 35.00 \\
MMLU Pro & Mistral & 49.00 & 51.00 & 51.00 \\
MMLU Pro & Qwen & 61.00 & 39.00 & 39.00 \\
MMLU Pro & Qwen-FT & 69.00 & 31.00 & 31.00 \\
QASC & Granite & 67.00 & 33.00 & 33.00 \\
QASC & Llama & 74.00 & 26.00 & 26.00 \\
QASC & Llama-FT & 77.00 & 23.00 & 23.00 \\
QASC & Mistral & 62.00 & 38.00 & 38.00 \\
QASC & Qwen & 72.00 & 28.00 & 28.00 \\
QASC & Qwen-FT & 80.00 & 20.00 & 20.00 \\
SuperGPQA & Granite & 41.00 & 59.00 & 59.00 \\
SuperGPQA & Llama & 32.00 & 68.00 & 68.00 \\
SuperGPQA & Llama-FT & 40.00 & 60.00 & 60.00 \\
SuperGPQA & Mistral & 37.00 & 63.00 & 63.00 \\
SuperGPQA & Qwen & 41.00 & 59.00 & 59.00 \\
SuperGPQA & Qwen-FT & 42.00 & 58.00 & 58.00 \\
\midrule
\textit{Avg} & \textit{All} & 62.54 $\pm$ 17.34 & 37.46 $\pm$ 17.34 & 37.46 $\pm$ 17.34 \\
\bottomrule
\end{tabular}
\end{table*}

\begin{table*}[!h]
\centering
\small
\caption{Augmented per-(dataset, model) results across all models (base and fine-tuned). Each cell shows the metric value with percentage change vs Original.}
\label{tab:appendix_augmented}
\begin{tabular}{llrrr}
\toprule
Dataset & Model & Precision $\uparrow$ & Error Rate $\downarrow$ & Composite Risk $\downarrow$ \\
\midrule
CommonsenseQA & Granite & 74.75 (+2.75) & 25.00 (-3.00) & 25.00 (-3.00) \\
CommonsenseQA & Llama & 78.00 (-2.00) & 22.00 (+2.00) & 22.00 (+2.00) \\
CommonsenseQA & Llama-FT & 87.00 (+1.00) & 13.00 (-1.00) & 13.00 (-1.00) \\
CommonsenseQA & Mistral & 75.76 (-0.24) & 24.00 (+0.00) & 24.00 (+0.00) \\
CommonsenseQA & Qwen & 85.00 (-1.00) & 15.00 (+1.00) & 15.00 (+1.00) \\
CommonsenseQA & Qwen-FT & 85.71 (-0.29) & 14.00 (+0.00) & 14.00 (+0.00) \\
MMLU Pro & Granite & 52.13 (+5.13) & 45.00 (-8.00) & 45.00 (-8.00) \\
MMLU Pro & Llama & 61.62 (+2.62) & 38.00 (-3.00) & 38.00 (-3.00) \\
MMLU Pro & Llama-FT & 69.00 (+4.00) & 31.00 (-4.00) & 31.00 (-4.00) \\
MMLU Pro & Mistral & 51.06 (+2.06) & 46.00 (-5.00) & 46.00 (-5.00) \\
MMLU Pro & Qwen & 66.33 (+5.33) & 33.00 (-6.00) & 33.00 (-6.00) \\
MMLU Pro & Qwen-FT & 69.00 (+0.00) & 31.00 (+0.00) & 31.00 (+0.00) \\
QASC & Granite & 69.07 (+2.07) & 30.00 (-3.00) & 30.00 (-3.00) \\
QASC & Llama & 73.00 (-1.00) & 27.00 (+1.00) & 27.00 (+1.00) \\
QASC & Llama-FT & 84.00 (+7.00) & 16.00 (-7.00) & 16.00 (-7.00) \\
QASC & Mistral & 68.69 (+6.69) & 31.00 (-7.00) & 31.00 (-7.00) \\
QASC & Qwen & 71.72 (-0.28) & 28.00 (+0.00) & 28.00 (+0.00) \\
QASC & Qwen-FT & 78.79 (-1.21) & 21.00 (+1.00) & 21.00 (+1.00) \\
SuperGPQA & Granite & 35.79 (-5.21) & 61.00 (+2.00) & 61.00 (+2.00) \\
SuperGPQA & Llama & 36.00 (+4.00) & 64.00 (-4.00) & 64.00 (-4.00) \\
SuperGPQA & Llama-FT & 48.00 (+8.00) & 52.00 (-8.00) & 52.00 (-8.00) \\
SuperGPQA & Mistral & 31.25 (-5.75) & 66.00 (+3.00) & 66.00 (+3.00) \\
SuperGPQA & Qwen & 46.46 (+5.46) & 53.00 (-6.00) & 53.00 (-6.00) \\
SuperGPQA & Qwen-FT & 40.00 (-2.00) & 57.00 (-1.00) & 57.00 (-1.00) \\
\midrule
\textit{Avg} & \textit{All} & 64.09 $\pm$ 17.27 (+1.55) & 35.12 $\pm$ 16.51 (-2.33) & 35.12 $\pm$ 16.51 (-2.33) \\
\bottomrule
\end{tabular}
\end{table*}

\begin{table*}[!h]
\centering
\small
\caption{Self-Evaluation per-(dataset, model) results across all models (base and fine-tuned). Each cell shows the metric value with percentage change vs Original.}
\label{tab:appendix_self_evaluation}
\begin{tabular}{llrrr}
\toprule
Dataset & Model & Precision $\uparrow$ & Error Rate $\downarrow$ & Composite Risk $\downarrow$ \\
\midrule
CommonsenseQA & Granite & 72.92 (+0.92) & 26.00 (-2.00) & 26.00 (-2.00) \\
CommonsenseQA & Llama & 86.57 (+6.57) & 9.00 (-11.00) & 9.00 (-11.00) \\
CommonsenseQA & Llama-FT & 86.87 (+0.87) & 13.00 (-1.00) & 13.00 (-1.00) \\
CommonsenseQA & Mistral & 78.00 (+2.00) & 22.00 (-2.00) & 22.00 (-2.00) \\
CommonsenseQA & Qwen & 86.87 (+0.87) & 13.00 (-1.00) & 13.00 (-1.00) \\
CommonsenseQA & Qwen-FT & 87.37 (+1.37) & 12.00 (-2.00) & 12.00 (-2.00) \\
MMLU Pro & Granite & 49.28 (+2.28) & 35.00 (-18.00) & 35.00 (-18.00) \\
MMLU Pro & Llama & 70.97 (+11.97) & 18.00 (-23.00) & 18.00 (-23.00) \\
MMLU Pro & Llama-FT & 65.00 (+0.00) & 35.00 (+0.00) & 35.00 (+0.00) \\
MMLU Pro & Mistral & 48.00 (-1.00) & 52.00 (+1.00) & 52.00 (+1.00) \\
MMLU Pro & Qwen & 64.89 (+3.89) & 33.00 (-6.00) & 33.00 (-6.00) \\
MMLU Pro & Qwen-FT & 70.10 (+1.10) & 29.00 (-2.00) & 29.00 (-2.00) \\
QASC & Granite & 67.39 (+0.39) & 30.00 (-3.00) & 30.00 (-3.00) \\
QASC & Llama & 74.67 (+0.67) & 19.00 (-7.00) & 19.00 (-7.00) \\
QASC & Llama-FT & 77.00 (+0.00) & 23.00 (+0.00) & 23.00 (+0.00) \\
QASC & Mistral & 64.00 (+2.00) & 36.00 (-2.00) & 36.00 (-2.00) \\
QASC & Qwen & 73.47 (+1.47) & 26.00 (-2.00) & 26.00 (-2.00) \\
QASC & Qwen-FT & 78.95 (-1.05) & 20.00 (+0.00) & 20.00 (+0.00) \\
SuperGPQA & Granite & 39.73 (-1.27) & 44.00 (-15.00) & 44.00 (-15.00) \\
SuperGPQA & Llama & 35.82 (+3.82) & 43.00 (-25.00) & 43.00 (-25.00) \\
SuperGPQA & Llama-FT & 39.39 (-0.61) & 60.00 (+0.00) & 60.00 (+0.00) \\
SuperGPQA & Mistral & 37.37 (+0.37) & 62.00 (-1.00) & 62.00 (-1.00) \\
SuperGPQA & Qwen & 43.62 (+2.62) & 53.00 (-6.00) & 53.00 (-6.00) \\
SuperGPQA & Qwen-FT & 100.00 (+58.00) & 0.00 (-58.00) & 0.00 (-58.00) \\
\midrule
\textit{Avg} & \textit{All} & 66.59 $\pm$ 18.41 (+4.05) & 29.71 $\pm$ 16.35 (-7.75) & 29.71 $\pm$ 16.35 (-7.75) \\
\bottomrule
\end{tabular}
\end{table*}

\begin{table*}[!h]
\centering
\small
\caption{Entropy Thresholding (Original) per-(dataset, model) results across all models (base and fine-tuned). Each cell shows the metric value with percentage change vs Original.}
\label{tab:appendix_entropy_thresholding_(original)}
\begin{tabular}{llrrr}
\toprule
Dataset & Model & Precision $\uparrow$ & Error Rate $\downarrow$ & Composite Risk $\downarrow$ \\
\midrule
CommonsenseQA & Granite & 71.72 (-0.28) & 28.00 (+0.00) & 29.00 (+1.00) \\
CommonsenseQA & Llama & 90.00 (+10.00) & 8.00 (-12.00) & 16.00 (-4.00) \\
CommonsenseQA & Llama-FT & 89.02 (+3.02) & 9.00 (-5.00) & 22.00 (+8.00) \\
CommonsenseQA & Mistral & 84.71 (+8.71) & 13.00 (-11.00) & 17.00 (-7.00) \\
CommonsenseQA & Qwen & 88.54 (+2.54) & 11.00 (-3.00) & 12.00 (-2.00) \\
CommonsenseQA & Qwen-FT & 94.74 (+8.74) & 4.00 (-10.00) & 18.00 (+4.00) \\
MMLU Pro & Granite & 47.87 (+0.87) & 49.00 (-4.00) & 51.00 (-2.00) \\
MMLU Pro & Llama & 61.25 (+2.25) & 31.00 (-10.00) & 41.00 (+0.00) \\
MMLU Pro & Llama-FT & 69.77 (+4.77) & 26.00 (-9.00) & 31.00 (-4.00) \\
MMLU Pro & Mistral & 52.44 (+3.44) & 39.00 (-12.00) & 45.00 (-6.00) \\
MMLU Pro & Qwen & 65.06 (+4.06) & 29.00 (-10.00) & 36.00 (-3.00) \\
MMLU Pro & Qwen-FT & 79.49 (+10.49) & 16.00 (-15.00) & 23.00 (-8.00) \\
QASC & Granite & 68.75 (+1.75) & 30.00 (-3.00) & 31.00 (-2.00) \\
QASC & Llama & 78.75 (+4.75) & 17.00 (-9.00) & 28.00 (+2.00) \\
QASC & Llama-FT & 84.52 (+7.52) & 13.00 (-10.00) & 19.00 (-4.00) \\
QASC & Mistral & 65.52 (+3.52) & 30.00 (-8.00) & 35.00 (-3.00) \\
QASC & Qwen & 76.40 (+4.40) & 21.00 (-7.00) & 25.00 (-3.00) \\
QASC & Qwen-FT & 87.65 (+7.65) & 10.00 (-10.00) & 19.00 (-1.00) \\
SuperGPQA & Granite & 43.96 (+2.96) & 51.00 (-8.00) & 52.00 (-7.00) \\
SuperGPQA & Llama & 31.96 (-0.04) & 66.00 (-2.00) & 67.00 (-1.00) \\
SuperGPQA & Llama-FT & 40.00 (+0.00) & 60.00 (+0.00) & 60.00 (+0.00) \\
SuperGPQA & Mistral & 41.56 (+4.56) & 45.00 (-18.00) & 50.00 (-13.00) \\
SuperGPQA & Qwen & 44.30 (+3.30) & 44.00 (-15.00) & 50.00 (-9.00) \\
SuperGPQA & Qwen-FT & 42.00 (+0.00) & 58.00 (+0.00) & 58.00 (+0.00) \\
\midrule
\textit{Avg} & \textit{All} & 66.67 $\pm$ 19.34 (+4.12) & 29.50 $\pm$ 18.28 (-7.96) & 34.79 $\pm$ 15.95 (-2.67) \\
\bottomrule
\end{tabular}
\end{table*}

\begin{table*}[!h]
\centering
\small
\caption{Entropy Thresholding (Augmented) per-(dataset, model) results across all models (base and fine-tuned). Each cell shows the metric value with percentage change vs Original.}
\label{tab:appendix_entropy_thresholding_(augmented)}
\begin{tabular}{llrrr}
\toprule
Dataset & Model & Precision $\uparrow$ & Error Rate $\downarrow$ & Composite Risk $\downarrow$ \\
\midrule
CommonsenseQA & Granite & 74.49 (+2.49) & 24.00 (-4.00) & 25.00 (-3.00) \\
CommonsenseQA & Llama & 85.19 (+5.19) & 12.00 (-8.00) & 21.00 (+1.00) \\
CommonsenseQA & Llama-FT & 88.89 (+2.89) & 9.00 (-5.00) & 24.00 (+10.00) \\
CommonsenseQA & Mistral & 79.76 (+3.76) & 17.00 (-7.00) & 25.00 (+1.00) \\
CommonsenseQA & Qwen & 87.37 (+1.37) & 12.00 (-2.00) & 14.00 (+0.00) \\
CommonsenseQA & Qwen-FT & 90.00 (+4.00) & 8.00 (-6.00) & 20.00 (+6.00) \\
MMLU Pro & Granite & 48.35 (+1.35) & 42.00 (-11.00) & 47.00 (-6.00) \\
MMLU Pro & Llama & 69.23 (+10.23) & 23.00 (-18.00) & 30.00 (-11.00) \\
MMLU Pro & Llama-FT & 72.41 (+7.41) & 24.00 (-11.00) & 30.00 (-5.00) \\
MMLU Pro & Mistral & 50.00 (+1.00) & 34.00 (-17.00) & 42.00 (-9.00) \\
MMLU Pro & Qwen & 72.09 (+11.09) & 22.00 (-17.00) & 25.00 (-14.00) \\
MMLU Pro & Qwen-FT & 78.21 (+9.21) & 17.00 (-14.00) & 25.00 (-6.00) \\
QASC & Granite & 69.07 (+2.07) & 28.00 (-5.00) & 28.00 (-5.00) \\
QASC & Llama & 80.49 (+6.49) & 16.00 (-10.00) & 23.00 (-3.00) \\
QASC & Llama-FT & 89.16 (+12.16) & 9.00 (-14.00) & 19.00 (-4.00) \\
QASC & Mistral & 71.43 (+9.43) & 24.00 (-14.00) & 32.00 (-6.00) \\
QASC & Qwen & 75.56 (+3.56) & 22.00 (-6.00) & 25.00 (-3.00) \\
QASC & Qwen-FT & 87.01 (+7.01) & 10.00 (-10.00) & 21.00 (+1.00) \\
SuperGPQA & Granite & 36.26 (-4.74) & 53.00 (-6.00) & 54.00 (-5.00) \\
SuperGPQA & Llama & 36.56 (+4.56) & 59.00 (-9.00) & 61.00 (-7.00) \\
SuperGPQA & Llama-FT & 50.55 (+10.55) & 45.00 (-15.00) & 47.00 (-13.00) \\
SuperGPQA & Mistral & 32.47 (-4.53) & 49.00 (-14.00) & 54.00 (-9.00) \\
SuperGPQA & Qwen & 47.50 (+6.50) & 41.00 (-18.00) & 49.00 (-10.00) \\
SuperGPQA & Qwen-FT & 37.37 (-4.63) & 57.00 (-1.00) & 58.00 (+0.00) \\
\midrule
\textit{Avg} & \textit{All} & 67.06 $\pm$ 19.29 (+4.52) & 27.38 $\pm$ 16.20 (-10.08) & 33.29 $\pm$ 14.11 (-4.17) \\
\bottomrule
\end{tabular}
\end{table*}

\begin{table*}[!h]
\centering
\small
\caption{Second Guess (Ours*) per-(dataset, model) results across all models (base and fine-tuned). Each cell shows the metric value wih Wpercentage change vs Original.}
\label{tab:detailed_second_guess_(ours*)}
\begin{tabular}{llrrr}
\toprule
Dataset & Model & Precision $\uparrow$ & Error Rate $\downarrow$ & Composite Risk $\downarrow$ \\
\midrule
CommonsenseQA & Granite & 81.82 (+9.82) & 14.00 (-14.00) & 23.00 (-5.00) \\
CommonsenseQA & Llama & 83.15 (+3.15) & 15.00 (-5.00) & 21.00 (+1.00) \\
CommonsenseQA & Llama-FT & 89.36 (+3.36) & 10.00 (-4.00) & 12.00 (-2.00) \\
CommonsenseQA & Mistral & 83.33 (+7.33) & 14.00 (-10.00) & 20.00 (-4.00) \\
CommonsenseQA & Qwen & 88.30 (+2.30) & 11.00 (-3.00) & 14.00 (+0.00) \\
CommonsenseQA & Qwen-FT & 88.17 (+2.17) & 11.00 (-3.00) & 15.00 (+1.00) \\
MMLU Pro & Granite & 63.79 (+16.79) & 21.00 (-32.00) & 31.00 (-22.00) \\
MMLU Pro & Llama & 72.06 (+13.06) & 19.00 (-22.00) & 29.00 (-12.00) \\
MMLU Pro & Llama-FT & 78.38 (+13.38) & 16.00 (-19.00) & 23.00 (-12.00) \\
MMLU Pro & Mistral & 58.46 (+9.46) & 27.00 (-24.00) & 38.00 (-13.00) \\
MMLU Pro & Qwen & 77.14 (+16.14) & 16.00 (-23.00) & 23.00 (-16.00) \\
MMLU Pro & Qwen-FT & 80.00 (+11.00) & 15.00 (-16.00) & 24.00 (-7.00) \\
QASC & Granite & 73.17 (+6.17) & 22.00 (-11.00) & 29.00 (-4.00) \\
QASC & Llama & 76.92 (+2.92) & 21.00 (-5.00) & 25.00 (-1.00) \\
QASC & Llama-FT & 84.62 (+7.62) & 14.00 (-9.00) & 14.00 (-9.00) \\
QASC & Mistral & 72.37 (+10.37) & 21.00 (-17.00) & 28.00 (-10.00) \\
QASC & Qwen & 76.47 (+4.47) & 20.00 (-8.00) & 27.00 (-1.00) \\
QASC & Qwen-FT & 81.72 (+1.72) & 17.00 (-3.00) & 21.00 (+1.00) \\
SuperGPQA & Granite & 46.51 (+5.51) & 23.00 (-36.00) & 44.00 (-15.00) \\
SuperGPQA & Llama & 40.74 (+8.74) & 32.00 (-36.00) & 42.00 (-26.00) \\
SuperGPQA & Llama-FT & 50.00 (+10.00) & 29.00 (-31.00) & 40.00 (-20.00) \\
SuperGPQA & Mistral & 46.34 (+9.34) & 22.00 (-41.00) & 40.00 (-23.00) \\
SuperGPQA & Qwen & 53.85 (+12.85) & 24.00 (-35.00) & 37.00 (-22.00) \\
SuperGPQA & Qwen-FT & 50.00 (+8.00) & 27.00 (-31.00) & 42.00 (-16.00) \\
\midrule
\textit{Avg} & \textit{All} & 70.69 $\pm$ 15.32 (+8.15) & 19.21 $\pm$ 5.92 (-18.25) & 27.58 $\pm$ 9.79 (-9.88) \\
\bottomrule
\end{tabular}
\end{table*}

\clearpage
\section{Second Guess: Change Breakdown for Remaining (Dataset, Model) Pairs}\label{sec:change_breakdown_combined}
Table~\ref{tab:change_breakdown_combined} shows how the responses given my models vary between the Original and Augmented prompt in the Second Guess technique, for all models and datasets evaluated.

\begin{table*}[h]
\centering
\small
\caption{Breakdown of paired-question outcomes under Second Guess across all remaining (dataset, model) pairs. Columns split by correctness on the Original prompt; within each group, the model's Augmented-prompt answer is one of: IDK, a different non-IDK option, or unchanged from the Original answer (``Preserved''). The two non-Preserved columns are forced to IDK by Second Guess. Counts are out of the paired-question set.}
\label{tab:change_breakdown_combined}
\begin{tabular}{llrrrrrrrr}
\toprule
 & & \multicolumn{4}{c}{Originally Correct} & \multicolumn{4}{c}{Originally Incorrect} \\
\cmidrule(lr){3-6} \cmidrule(lr){7-10}
Dataset & Model & $\rightarrow$ IDK & $\rightarrow$ Other & Preserved & Total & $\rightarrow$ IDK & $\rightarrow$ Other & Preserved & Total \\
\midrule
CommonsenseQA & Granite & 0 & 9 & 63 & 72 & 1 & 13 & 14 & 28 \\
CommonsenseQA & Llama & 0 & 6 & 74 & 80 & 0 & 5 & 15 & 20 \\
CommonsenseQA & Llama-FT & 0 & 2 & 84 & 86 & 0 & 4 & 10 & 14 \\
CommonsenseQA & Mistral & 0 & 6 & 70 & 76 & 1 & 9 & 14 & 24 \\
CommonsenseQA & Qwen & 0 & 3 & 83 & 86 & 0 & 3 & 11 & 14 \\
CommonsenseQA & Qwen-FT & 2 & 2 & 82 & 86 & 0 & 3 & 11 & 14 \\
MMLU Pro & Granite & 2 & 8 & 37 & 47 & 4 & 28 & 21 & 53 \\
MMLU Pro & Llama & 0 & 10 & 49 & 59 & 1 & 21 & 19 & 41 \\
MMLU Pro & Llama-FT & 0 & 7 & 58 & 65 & 0 & 19 & 16 & 35 \\
MMLU Pro & Mistral & 1 & 10 & 38 & 49 & 5 & 19 & 27 & 51 \\
MMLU Pro & Qwen & 0 & 7 & 54 & 61 & 2 & 21 & 16 & 39 \\
MMLU Pro & Qwen-FT & 0 & 9 & 60 & 69 & 0 & 16 & 15 & 31 \\
QASC & Granite & 1 & 6 & 60 & 67 & 2 & 9 & 22 & 33 \\
QASC & Llama & 0 & 4 & 70 & 74 & 0 & 5 & 21 & 26 \\
QASC & Llama-FT & 0 & 0 & 77 & 77 & 0 & 9 & 14 & 23 \\
QASC & Mistral & 1 & 6 & 55 & 62 & 0 & 17 & 21 & 38 \\
QASC & Qwen & 1 & 6 & 65 & 72 & 0 & 8 & 20 & 28 \\
QASC & Qwen-FT & 1 & 3 & 76 & 80 & 0 & 3 & 17 & 20 \\
SuperGPQA & Granite & 2 & 19 & 20 & 41 & 3 & 33 & 23 & 59 \\
SuperGPQA & Llama & 0 & 10 & 22 & 32 & 0 & 36 & 32 & 68 \\
SuperGPQA & Llama-FT & 0 & 11 & 29 & 40 & 0 & 31 & 29 & 60 \\
SuperGPQA & Mistral & 3 & 15 & 19 & 37 & 1 & 40 & 22 & 63 \\
SuperGPQA & Qwen & 0 & 13 & 28 & 41 & 1 & 34 & 24 & 59 \\
SuperGPQA & Qwen-FT & 2 & 13 & 27 & 42 & 3 & 28 & 27 & 58 \\
\bottomrule
\end{tabular}
\end{table*}

\end{document}